\documentclass[sigconf]{acmart}

\usepackage{booktabs} 

\setcopyright{rightsretained}

\acmDOI{10.475/123_4}

\acmISBN{123-4567-24-567/08/06}

\acmConference[GECCO '17]{the Genetic and Evolutionary Computation
  Conference 2017}{July 15--19, 2017}{Berlin, Germany}
\acmYear{2017}
\copyrightyear{2017}

\acmPrice{15.00}

\begin{document}
\title{Evolving Deep Neural Networks}

\author{Risto Miikkulainen$^{1,2}$, Jason Liang$^{1,2}$, Elliot
  Meyerson$^{1,2}$, Aditya Rawal$^{1,2}$, Dan Fink$^{1}$, Olivier
  Francon$^{1}$, Bala Raju$^{1}$, Hormoz Shahrzad$^{1}$, Arshak
  Navruzyan$^{1}$, Nigel Duffy$^{1}$, Babak Hodjat$^{1}$\\
\mbox{}\\
$^1$Sentient Technologies, Inc.\\
$^2$The University of Texas at Austin\\
}

\begin{abstract}

The success of deep learning depends on finding an architecture to
fit the task.  As deep learning has scaled up to more challenging
tasks, the architectures have become difficult to design by
hand. This paper proposes an automated method, CoDeepNEAT, for
optimizing deep learning architectures through evolution. By
extending existing neuroevolution methods to topology, components,
and hyperparameters, this method achieves results comparable to best
human designs in standard benchmarks in object recognition and
language modeling. It also supports building a real-world
application of automated image captioning on a magazine
website. Given the anticipated increases in available computing
power, evolution of deep networks is promising approach to
constructing deep learning applications in the future.

\end{abstract}

%
%

\ccsdesc[500]{Computing methodologies~Genetic algorithms}
\ccsdesc[500]{Computing methodologies~Neural networks}

%
\keywords{Neural networks, deep learning, LSTMs, bilevel optimization,
  coevolution, design}

\maketitle

\section{Introduction}
\label{sec:intro}

Large databases (i.e.\ Big Data) and large amounts of computing power
have become readily available since the 2000s. As a result, it has
become possible to scale up machine learning systems. Interestingly,
not only have these systems been successful in such scaleup, but they
have become more powerful. Some ideas that did not quite work before,
now do, with million times more compute and data. For instance, deep
learning neural networks (DNNs), i.e.\ convolutional neural networks
\cite{lecun:ieee98} and recurrent neural networks (in particular long
short-term memory, or LSTM \cite{hochreiter:nc97}), which have existed
since the 1990s, have improved state-of-the-art significantly in
computer vision, speech, language processing, and many other areas
\cite{collobert:icml08,graves:icassp13,szegedy:cvpr16}.

As DNNs have been scaled up and improved, they have become much more
complex.  A new challenge has therefore emerged: How to configure such
systems? Human engineers can optimize a handful of configuration
parameters through experimentation, but DNNs have complex topologies
and hundreds of hyperparameters.  Moreover, such design choices
matter; often success depends on finding the right architecture for
the problem. Much of the recent work in deep learning has indeed
focused on proposing different hand-designed architectures on new
problems \cite{he:arxiv16,ng:arxiv15,che:arxiv16,szegedy:cvpr16}.

The complexity challenge is not unique to neural networks. Software
and many other engineered systems have become too complex for humans
to optimize fully. As a result, a new way of thinking about such
design has started to emerge. In this approach, humans are responsible
for the high-level design, and the details are left for computational
optimization systems to figure out. For instance, humans write the
overall design of a software system, and the parameters and low-level
code is optimized automatically \cite{hoos:acm12}; humans write
imperfect versions of programs, and evolutionary algorithms are then
used to repair them \cite{goues:softeng12}; humans define the space of
possible web designs, and evolution is used to find effective ones
\cite{miikkulainen:gecco17-ascend}.

This same approach can be applied to the design of DNN
architectures. This problem includes three challenges: how to design
the components of the architecture, how to put them together into a
full network topology, and how to set the hyperparameters for the
components and the global design.  These three aspects need to be
optimized separately for each new task.

This paper develops an approach for automatic design of DNNs. It is
based on the existing neuroevolution technique of NEAT
\cite{stanley:ec02}, which has been successful in evolving topologies
and weights of relatively small recurrent networks in the past. In
this paper, NEAT is extended to the coevolutionary optimization of
components, topologies, and hyperparameters. The fitness of the
evolved networks is determined based on how well they can be trained,
through gradient descent, to perform in the task. The approach is
demonstrated in the standard benchmark tasks of object recognition and
language modeling, and in a real-world application of captioning
images on a magazine website.

The results show that the approach discovers designs that are
comparable to the state of the art, and does it automatically without
much development effort. The approach is computationally extremely
demanding---with more computational power, it is likely to be more
effective and possibly surpass human design. Such power is now
becoming available in various forms of cloud computing and grid
computing, thereby making evolutionary optimization of neural networks
a promising approach for the future.

\section{Background and Related Work}
\label{sec:bg}

Neuroevolution techniques have been applied successfully to sequential
decision tasks for three decades
\cite{montana:ijcai89,floreano:evolint08,yao:ieee99,lehman:scholarpedia13}.
In such tasks there is no gradient available, so instead of gradient
descent, evolution is used to optimize the weights of the neural
network. Neuroevolution is a good approach in particular to POMDP
(partially observable Markov decision process) problems because of
recurrency: It is possible to evolve recurrent connections to allow
disambiguating hidden states.

The weights can be optimized using various evolutionary techniques.
Genetic algorithms are a natural choice because crossover is a good
match with neural networks: they recombine parts of existing neural
networks to find better ones. CMA-ES \cite{igel:ieeecec03}, a technique
for continuous optimization, works well on optimizing the weights as
well because it can capture interactions between them. Other
approaches such as SANE, ESP, and CoSyNE evolve partial neural
networks and combine them into fully functional networks
\cite{moriarty:ec97,gomez:ab97,gomez:jmlr08}.  Further, techniques
such as Cellular Encoding \cite{gruau:baldwin} and NEAT
\cite{stanley:ec02} have been developed to evolve the topology of the
neural network, which is particularly effective in determining the
required recurrence.  Neuroevolution techniques have been shown to
work well in many tasks in control, robotics, constructing intelligent
agents for games, and artificial life
\cite{lehman:scholarpedia13}. However, because of the large number of
weights to be optimized, they are generally limited to relatively
small networks.

Evolution has been combined with gradient-descent based learning in
several ways, making it possible to utilize much larger networks.
These methods are still usually applied to sequential decision tasks,
but gradients from a related task (such as prediction of the next
sensory inputs) are used to help search. Much of the work is based on
utilizing the Baldwin effect, where learning only affects the
selection \cite{hinton:evolution}. Computationally, it is possible to
utilize Lamarckian evolution as well, i.e.\ encode the learned weight
changes back into the genome \cite{gruau:baldwin}. However, care must
be taken to maintain diversity so that evolution can continue to
innovate when all individuals are learning similar behavior.

Evolution of DNNs departs from this prior work in that it is applied
to supervised domains where gradients are available, and evolution is
used only to optimize the design of the neural network.  Deep
neuroevolution is thus more closely related to bilevel (or multilevel)
optimization techniques \cite{sinha:gecco14}. The idea is to use an
evolutionary optimization process at a high level to optimize the
parameters of a low-level evolutionary optimization process.

Consider for instance the problem of controlling a helicopter through
aileron, elevator, rudder, and rotor inputs. This is a challenging
benchmark from the 2000s for which various reinforcement learning
approaches have been developed
\cite{bagnell:icra01,ng:nips16,abbeel:nips07}.  One of the most
successful ones is single-level neuroevolution, where the helicopter
is controlled by a neural network that is evolved through genetic
algorithms \cite{koppejan:evolint11}. The eight parameters of the
neuroevolution method (such mutation and crossover rate, probability,
and amount and population and elite size) are optimized by hand.  It
would be difficult to include more parameters because the parameters
interact nonlinearly. A large part of the parameter space thus remains
unexplored in the single-level neuroevolution approach.  However, a
bilevel approach, where a high-level evolutionary process is employed
to optimize these parameters, can search this space more effectively
\cite{liang:gecco15}.  With bilevel evolution, the number of
parameters optimized could be extended to 15, which resulted in
significantly better performance. In this manner, evolution was
harnessed in this example task to optimize a system design that was
too complex to be optimized by hand.

Very recently, a studies have started to emerge with the goal of
optimizing DNNs. Due to limited computational resources, they have
focused on specific parts of the design. For instance, Loshchilov et
al.\ \cite{loshchilov:arxiv16} used CMA-ES to optimize the
hyperparameters of existing DNNs obtaining state-of-the-art results on
e.g.\ object recognition.  Further, Fernando et
al. \cite{fernando:gecco16} evolved a CPPN (compositional
pattern-producing network \cite{stanley:gpem07}) to output the weights
of an auto-encoder neural network. The autoencoder was then trained
further through gradient descent, forming gradients for the CPPN
training, and its trained weights were then incorporated back into the
CPPN genome through Lamarckian adaptation. A related approach was
proposed by Zoph and Le \cite{zoph:arxiv16}: the topology and
hyperparameters of a deep network and LSTM network were modified
through policy iteration.

Building on this foundation, a systematic approach to evolving DNNs is
developed in this paper. First, the standard NEAT neuroevolution
method is applied to the topology and hyperparameters of convolutional
neural networks, and then extended to evolution of components as well,
achieving results comparable to state of the art in the CIFAR-10 image
classification benchmark. Second, a similar method is used to evolve
the structure of LSTM networks in language modeling, showing that even
small innovations in the components can have a significant effect on
performance. Third, the approach is used to build a real-world
application on captioning images in the website for an online
magazine.

\begin{figure}[t]
  \begin{center}
    \includegraphics[width=\columnwidth]{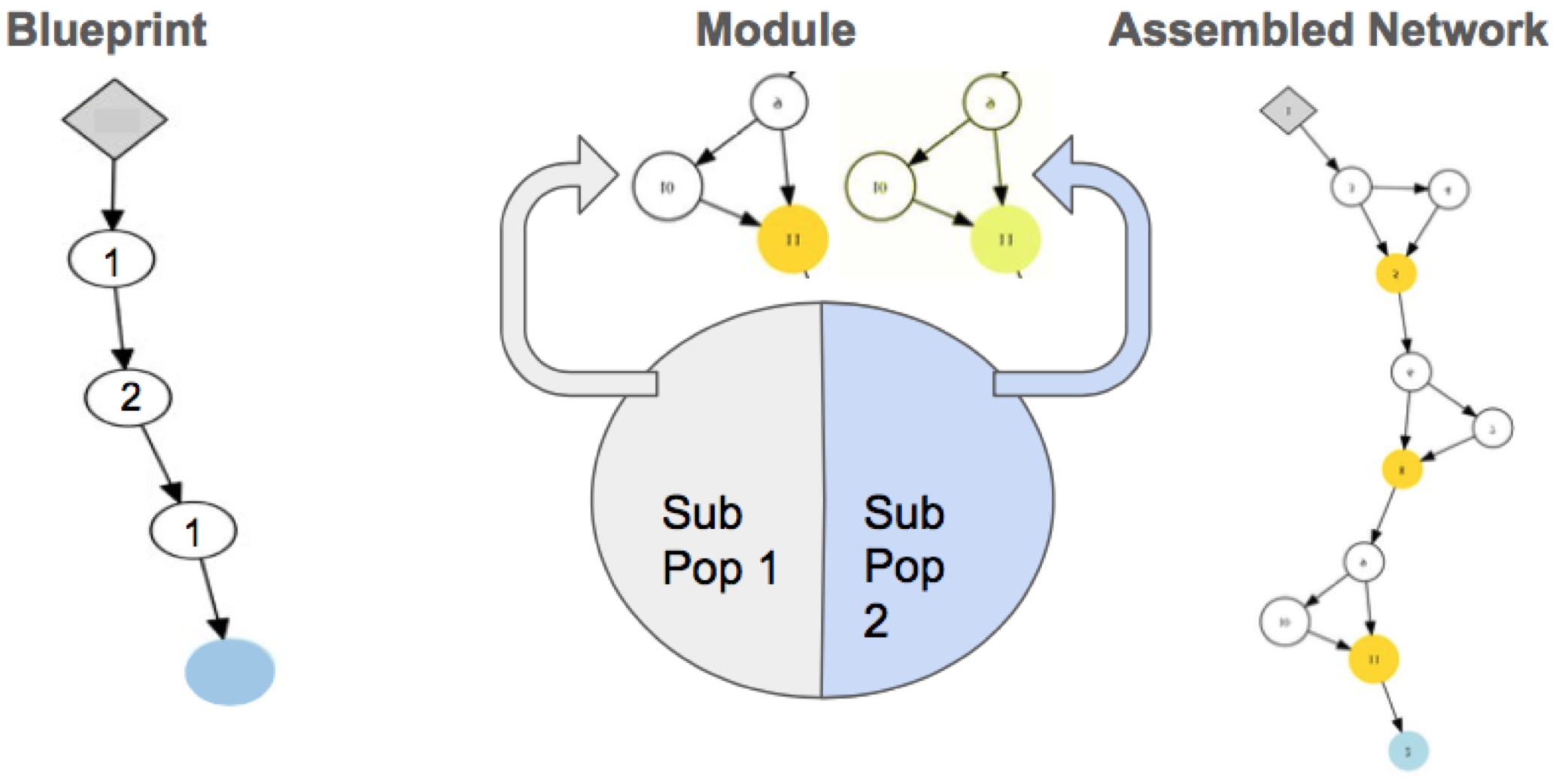}
    \caption{A visualization of how CoDeepNEAT assembles networks for
      fitness evaluation. Modules and blueprints are assembled
      together into a network through replacement of blueprint nodes
      with corresponding modules. This approach allows evolving
      repetitive and deep structures seen in many successful recent
      DNNs.}
    \label{fg:assembling}
  \end{center}
\end{figure}


\section{Evolution of Deep Learning Architectures}
\label{sec:dl}

NEAT neuroevolution method \cite{stanley:ec02} is first extended to
evolving network topology and hyperparameters of deep neural networks
in DeepNEAT, and then further to coevolution of modules and
blueprints for combining them in CoDeepNEAT.  The approach is tested
in the standard CIFAR-10 benchmark of object recognition, and found to
be comparable to the state of the art.

\subsection{Extending NEAT to Deep Networks}

DeepNEAT is a most immediate extension of the standard neural network
topology-evolution method NEAT to DNN. It follows the same fundamental
process as NEAT: First, a population of chromosomes (each represented
by a graph) with minimal complexity is created.  Over generations,
structure (i.e.\ nodes and edges) is added to the graph incrementally
through mutation. During crossover, historical markings are used to
determine how genes of two chromosomes can be lined up. The population
is divided into species (i.e.\ subpopulations) based on a similarity
metric.  Each species grows proportionally to its fitness and
evolution occurs separately in each species.

DeepNEAT differs from NEAT in that each node in the chromosome no
longer represents a neuron, but a layer in a DNN.  Each node contains
a table of real and binary valued hyperparameters that are mutated
through uniform Gaussian distribution and random bit-flipping,
respectively. These hyperparameters determine the type of layer (such
as convolutional, fully connected, or recurrent) and the properties of
that layer (such as number of neurons, kernel size, and activation
function). The edges in the chromosome are no longer marked with
weights; instead they simply indicate how the nodes (layers) are
connected together. To construct a DNN from a DeepNEAT chromosome, one
simply needs to traverse the chromosome graph, replacing each node
with the corresponding layer. The chromosome also contains a set of
global hyperparameters applicable to the entire network (such as
learning rate, training algorithm, and data preprocessing).

When arbitrary connectivity is allowed between layers, additional
complexity is required. If the current layer has multiple parent
layers, a merge layer must be applied to the parents in order to
ensure that the parent layer's output is the same size as the current
layer's input. Typically, this adjustment is done through a
concatenation or element-wise sum operation. If the parent layers have
mismatched output sizes, all of the parent layers must be downsampled
to parent layer with the smallest output size. The specific method for
downsampling is domain dependent. For example, in image
classification, a max-pooling layer is inserted after specific parent
layers; in image captioning, a fully connected bottleneck layer will
serve this function.

During fitness evaluation, each chromosome is converted into a
DNN. These DNNs are then trained for a fixed number of epochs. After
training, a metric that indicates the network's performance is
returned back to DeepNEAT and assigned as fitness to the corresponding
chromosome in the population.

While DeepNEAT can be used to evolve DNNs, the resulting structures
are often complex and unprincipled. They contrast with typical DNN
architectures that utilize repetition of basic components. DeepNEAT is
therefore extend to evolution of modules and blueprints next.



\subsection{Cooperative Coevolution of Modules and Blueprints}
\label{subsec:coevolution}

Many of the most successful DNNs, such as GoogLeNet and ResNet are
composed of modules that are repeated multiple times
\cite{szegedy:cvpr16,he:arxiv16}. These modules often themselves have
complicated structure with branching and merging of various
layers. Inspired by this observation, a variant of DeepNEAT, called
Coevolution DeepNEAT (CoDeepNEAT), is proposed. The algorithm behind
CoDeepNEAT is inspired mainly by Hierarchical SANE
\cite{moriarty:ec97} but is also influenced by component-evolution
approaches ESP \cite{gomez:ijcai99} and CoSyNE \cite{gomez:jmlr08}.

In CoDeepNEAT, two populations of modules and blueprints are evolved
separately, using the same methods as described above for
DeepNEAT. The blueprint chromosome is a graph where each node contains
a pointer to a particular module species.  In turn, each module
chromosome is a graph that represents a small DNN. During fitness
evaluation, the modules and blueprints are combined together to create
a larger assembled network Figure~\ref{fg:assembling}.  Each node in
the blueprint is replaced with a module chosen randomly from the
species to which that node points. If multiple blueprint nodes point
to the same module species, then the same module is used in all of
them. The assembled networks are evaluated the a manner similar to
DeepNEAT, but the fitnesses of the assembled networks are attributed
back to blueprints and modules as the average fitness of all the
assembled networks containing that blueprint or module.

CoDeepNEAT can evolve repetitive modular structure efficiently.
Furthermore, because small mutations in the modules and blueprints
often lead to large changes in the assembled network structure,
CoDeepNEAT can explore more diverse and deeper architectures than
DeepNEAT. An example application to the CIFAR-10 domain is presented
next.




\subsection{Evolving DNNs in the CIFAR-10 Benchmark}

\begin{table}[]
\centering
\begin{tabular}{l r} \hline
\textbf{Node Hyperparameter} & \textbf{Range} \\ \hline
Number of Filters & $[32, 256]$ \\
Dropout Rate & $[0, 0.7]$ \\
Initial Weight Scaling & $[0, 2.0]$ \\
Kernel Size & $\{1, 3\}$ \\
Max Pooling & \{True, False\} \\ \hline
\textbf{Global Hyperparameter} & \textbf{Range} \\ \hline
Learning Rate & $[0.0001, 0.1]$ \\
Momentum & $[0.68, 0.99]$ \\
Hue Shift & $[0, 45]$ \\
Saturation/Value Shift & $[0, 0.5]$ \\
Saturation/Value Scale & $[0, 0.5]$ \\
Cropped Image Size & $[26, 32]$ \\
Spatial Scaling & $[0, 0.3]$ \\
Random Horizontal Flips & \{True, False\} \\
Variance Normalization & \{True, False\} \\
Nesterov Accelerated Gradient & \{True, False\} \\
\end{tabular}
\caption{Node and global hyperparameters evolved in the CIFAR-10 domain.}
\label{tb:cifar10}
\end{table}

In this experiment, CoDeepNEAT was used to evolve the topology of a
convolutional neural network (CNN) to maximize its classification
performance on the CIFAR-10 dataset, a common image classification
benchmark. The dataset consists of 50,000 training images and 10,000
testing images. The images consist of 32x32 color pixels and belong to
one of 10 classes. For comparison, the neural network layer types were
restricted to those used by Snoek et al.\ \cite{snoek:icml15} in
their Bayesian optimization of CNN hyperparameters.  Also following
Snoek et al., data augmentation consisted of converting the images
from RGB to HSV color space, adding random perturbations, distortions,
and crops, and converting them back to RGB color space.

CoDeepNEAT was initialized with populations of 25 blueprints and 45
modules. From these two populations, 100 CNNs were assembled for
fitness evaluation in every generation.  Each node in the module
chromosome represents a convolutional layer. Its hyperparameters
determine the various properties of the layer and whether additional
max-pooling or dropout layers are attached (Table~\ref{tb:cifar10}).
In addition, a set of global hyperparameters were evolved for the
assembled network. During fitness evaluation, the 50,000 images were
split into a training set of 42,500 samples and a validation set of
7,500 samples.  Since training a DNN is computationally very
expensive, each network was trained for eight epochs on the training
set. The validation set was then used to determine classification
accuracy, i.e.\ the fitness of the network. After 72 generations of
evolution, the best network in the population was returned.

\begin{figure}[t]
  \begin{center}
    \includegraphics[width=0.45\textwidth]{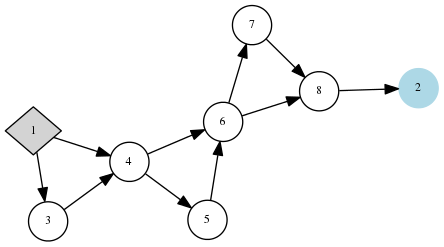}
    \includegraphics[width=0.45\textwidth]{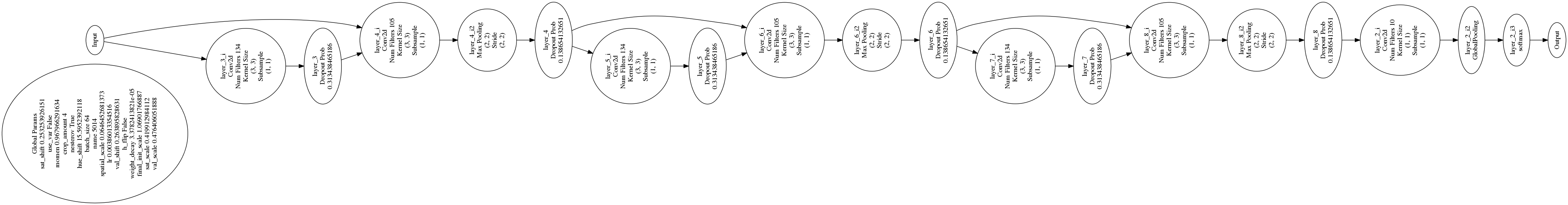}
    \caption{Top: Simplified visualization of the best network evolved
      by CoDeepNEAT for the CIFAR-10 domain. Node 1 is the input layer,
      while Node 2 is the output layer. The network has repetitive
      structure because its blueprint reuses same module in multiple
      places. Bottom: A more detailed visualization of the same
      network. The full image is included in supplementary material.}
    \label{fg:cifar10_1}
  \end{center}
\end{figure}

After evolution was complete, the best network was trained on all
50,000 training images for 300 epochs, and the classification error
measured.  This error was 7.3\%, comparable to the 6.4\% error
reported by Snoek et al.\ \cite{snoek:icml15}. Interestingly, because
only limited training could be done during evolution, the best network
evolved by CoDeepNEAT trains very fast. While the network of Snoek et
al.\ takes over 30 epochs to reach 20\% test error and over 200 epochs
to converge, the best network from evolution takes only 12 epochs to
reach 20\% test error and around 120 epochs to converge. This network
utilizes the same modules multiple times, resulting in a deep and
repetitive structure typical of many successful DNNs
(Figure~\ref{fg:cifar10_1}).

\section{Evolution of LSTM Architectures}

Recurrent neural networks, in particular those utilizing LSTM nodes,
is another powerful approach to DNN. Much of the power comes from
repetition of LSTM modules and the connectivity between them. In this
section, CoDeepNEAT is extended with mutations that allow searching
for such connectivity, and the approach is evaluated in the standard
benchmark task of language modeling.

\subsection{Extending CoDeepNEAT to LSTMs}

Long Short Term Memory (LSTM) consists of gated memory cells that can
integrate information over longer time scales (as compared to simply
using recurrent connections in a neural network). LSTMs have recently
been shown powerful in supervised sequence processing tasks such as
speech recognition \cite{graves:icml} and machine translation
\cite{bahdanau:iclr}.
  
Recent research on LSTMs has focused in two directions: Finding
variations of individual LSTM memory unit architecture
\cite{bayer:icann, jozefowicz:icml, cho:gru, klaus:arxiv14}, and
discovering new ways of stitching LSTM layers into a network
\cite{chung:arxiv15,kalchbrenner:gridlstm,zilly:highwaylstm}.  Both
approaches have improved performance over vanilla LSTMs, with best
recent results achieved through network design.  The CoDeepNEAT method
incorporates both approaches: neuroevolution searches for both new
LSTM units and their connectivity across multiple layers at the same
time.
  
CoDeepNEAT was slightly modified to make it easier to find novel
connectivities between LSTM layers. Multiple LSTM layers are flattened
into a neural network graph that is then modified by neuroevolution.
There are two types of mutations: one enables or disables a connection
between LSTM layers, and the other adds or removes skip connections
between two LSTM nodes. Recently, skip connections have led to
performance improvements in deep neural networks, which suggests that
they could be useful for LSTM networks as well. Thus, neuroevolution
modifies both the high-level network topology and the low-level LSTM
connections.

In each generation, a population of these network graphs (i.e.\
blueprints), consisting of LSTM variants (i.e.\ modules with possible
skip connections), is created. The individual networks are then
trained and tested with the supervised data of the task.  The
experimental setup and the language modeling task are described next.

\subsection{Evolving DNNs in the Language Modeling Benchmark}

One standard benchmark task for LSTM network is language modeling,
i.e.\ predicting the next word in a large text corpus. The benchmark
utilizes the Penn Tree Bank (PTB) dataset \cite {marcus:ptb}, which
consists of 929k training words, 73k validation words, and 82k test
words. It has 10k words in its vocabulary.

A population of 50 LSTM networks was initialized with uniformly random
initial connection weights within [-0.05, 0.05]. Each network
consisted of two recurrent layers (vanilla LSTM or its variants) with
650 hidden nodes in each layer. The network was unrolled in time upto
35 steps. The hidden states were initialized to zero. The final hidden
states of the current minibatch was used as the initial hidden state
of the subsequent minibatch (successive minibatches sequentially
traverse the training set). The size of each minibatch was 20. For
fitness evaluation, each network was trained for 39 epochs. A learning
rate decay of 0.8 was applied at the end of every six epochs; the
dropout rate was 0.5. The gradients were clipped if their maximum norm
(normalized by minibatch size) exceeded 5. Training a single network
took about 200 minutes on a GeForce GTX 980 GPU card.

\begin{figure}
  \begin{center}
    \includegraphics[width=0.5\columnwidth]{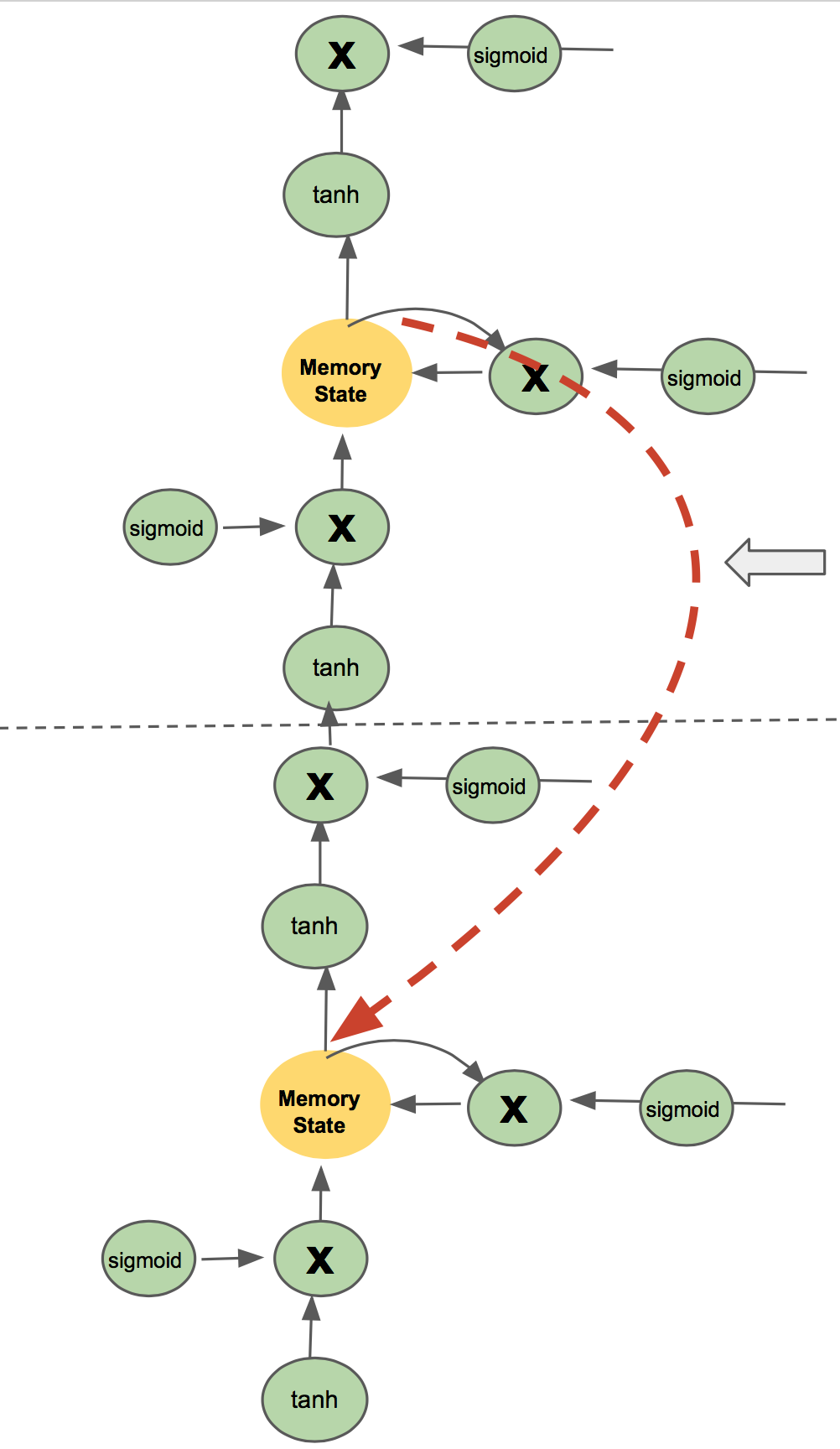}
    \caption{The best-performing LSTM variant after 25 generations of
      neuroevolution. It includes a novel skip connection between the
      two memory cells, resulting in 5\% improvement over the vanilla
      LSTM baseline. Such improvements are difficult to discover by
      hand; CoDeepNEAT with LSTM-specific mutation searches for them
      automatically.}
    \label{fg:lstm}
  \end{center}
\end{figure}

After 25 generations of neuroevolution, the best network improved the
performance on PTB dataset by 5\% (test-perplexity score 78) as
compared to the vanilla LSTM \cite{zaremba:arxiv14}. As shown in
Figure\ref{fg:lstm}, this LSTM variant consists of a feedback skip
connection between the memory cells of two LSTM layers. This result is
interesting because it is similar to a recent hand-designed
architecture that also outperforms vanilla LSTM \cite{chung:arxiv15}.

The initial results thus demonstrate that CoDeepNEAT with just two
LSTM-specific mutations can automatically discover improved LSTM variants.
It is likely that expanding the search space with more mutation types
and layer and connection types would lead to further improvements. 

\section{Application Case Study: Image Captioning for the Blind}

In a real-world case study, the vision and language capabilities of
CoDeepNEAT were combined to build a real-time online image captioning
system.  In this application, CoDeepNEAT searches for architectures
that learn to integrate image and text representations to produce
captions that blind users can access through existing screen readers.
This application was implemented for a major online magazine website.
Evolved networks
were trained with the open source MSCOCO image captioning dataset
\cite{chen:arxiv15}, along with a new dataset collected for this
website.

\subsection{Evolving DNNs for Image Captioning}

Deep learning has recently provided state-of-the-art performance in
image captioning, and several diverse architectures have been
suggested \cite{vinyals:cvpr15, xu:icml15, karpathy:cvpr15,
  you:cvpr16, vedantam:arxiv17}.  The input to an image captioning
system is a raw image, and the output is a text caption intended to
describe the contents of the image. In deep learning approaches, a
convolutional network is usually used to process the image, and
recurrent units, often LSTMs, to generate coherent sentences with
long-range dependencies.

As is common in existing approaches, the evolved system uses a
pretrained ImageNet model \cite{szegedy:cvpr16} to produce initial
image embeddings. The evolved network takes an image embedding as
input, along with a one-hot text input. As usual, in training the text
input contains the previous word of the ground truth caption; in
inference it contains the previous word generated by the model
\cite{vinyals:cvpr15, karpathy:cvpr15}.

In the initial CoDeepNEAT population the image and text inputs are fed
to a shared embedding layer, which is densely connected to a softmax
output over words. From this simple starting point, CoDeepNEAT evolves
architectures that include fully connected layers, LSTM layers, sum
layers, concatenation layers, and sets of hyperparameters associated
with each layer, along with a set of global hyperparameters
(Table~\ref{tb:img_cap_hypers}). In particular, the well-known Show
and Tell image captioning architecture \cite{vinyals:cvpr15} is in
this search space, providing a baseline with which evolution results
can be compared.
\begin{table}[]
\centering
\begin{tabular}{l r} \hline
\textbf{Global Hyperparameter} & \textbf{Range} \\\hline
Learning Rate & $[0.0001, 0.1]$ \\
Momentum & $[0.68, 0.99]$ \\
Shared Embedding Size &  $[128, 512]$ \\
Embedding Dropout & $[0, 0.7]$ \\
LSTM Recurrent Dropout & \{True, False\} \\
Nesterov Momentum & \{True, False\} \\
Weight Initialization & \{Glorot normal, He normal\} \\ \hline
\textbf{Node Hyperparameter} & \textbf{Range} \\ \hline
Layer Type & \{Dense, LSTM\} \\
Merge Method & \{Sum, Concat\} \\
Layer Size & $\{128, 256\}$ \\
Layer Activation & \{ReLU, Linear\} \\
Layer Dropout & $[0, 0.7]$ \\
\end{tabular}
\caption{Node and global hyperparameters evolved for the image
  captioning case study.}
\label{tb:img_cap_hypers}
\end{table}
These components and the glue that connects them are evolved as
described in Section~\ref{subsec:coevolution}, with 100 networks
trained in each generation. Since there is no single best accepted
metric for evaluating captions, the fitness function is the mean
across three metrics (BLEU, METEOR, and CIDEr; \cite{chen:arxiv15})
normalized by their baseline values. Fitness is computed over a
holdout set of 5000 images, i.e.\ 25,000 image-caption pairs.

To keep the computational cost reasonable, during evolution the
networks are trained for only six epochs, and only with a random
100,000 image subset of the 500,000 MSCOCO image-caption pairs. As a
result, there is evolutionary pressure towards networks that converge
quickly: The best resulting architectures train to near convergence 6
times faster than the baseline Show and Tell model
\cite{vinyals:cvpr15}.  After evolution, the optimized learning rate
is scaled by one-fifth to compensate for the subsampling.

\subsection{Building the Application}

The images in MSCOCO are chosen to depict ``common objects in
context''. The focus is on a relatively small set of objects and their
interactions in a relatively small set of settings. The internet as a
whole, and the online magazine website in particular, contain many
images that cannot be classified as ``common objects in
context''. Other types of images from the magazine include staged
portraits of people, infographics, cartoons, abstract designs, and
\emph{iconic} images, i.e.\ images of one or multiple objects
\emph{out of context} such as on a white or patterned background.
Therefore, an additional dataset of 17,000 image-caption pairs was
constructed for the case study, targeting iconic images in particular.
Four thousand images were first scraped from the magazine website, and
1000 of them were identified as iconic.  Then, 16,000 images that were
visually similar to those 1000 were retrieved automatically from a
large image repository. A single ground truth caption for each of
these 17K images was generated by human subjects through
Spare5\footnote{https://mty.ai/computer-vision/}. The holdout set for
evaluation consisted of 100 of the original 1000 iconic images, along
with 3000 other images.

During evolution, networks were trained and evaluated only on the
MSCOCO data. The best architecture from evolution was then trained
from scratch on both the MSCOCO and Spare5 datasets in an iterative
alternating approach: one epoch on MSCOCO, followed by five epochs on
Spare5, until maximum performance was reached on the Spare5 holdout
data. Beam search was then used to generate captions from the fully
trained models. Performance achieved using the Spare5 data
demonstrates the ability of evolved architectures to generalize to
domains towards which they were not evolved.

\begin{figure}[]
  \begin{center}
    \includegraphics[width=0.4\textwidth]{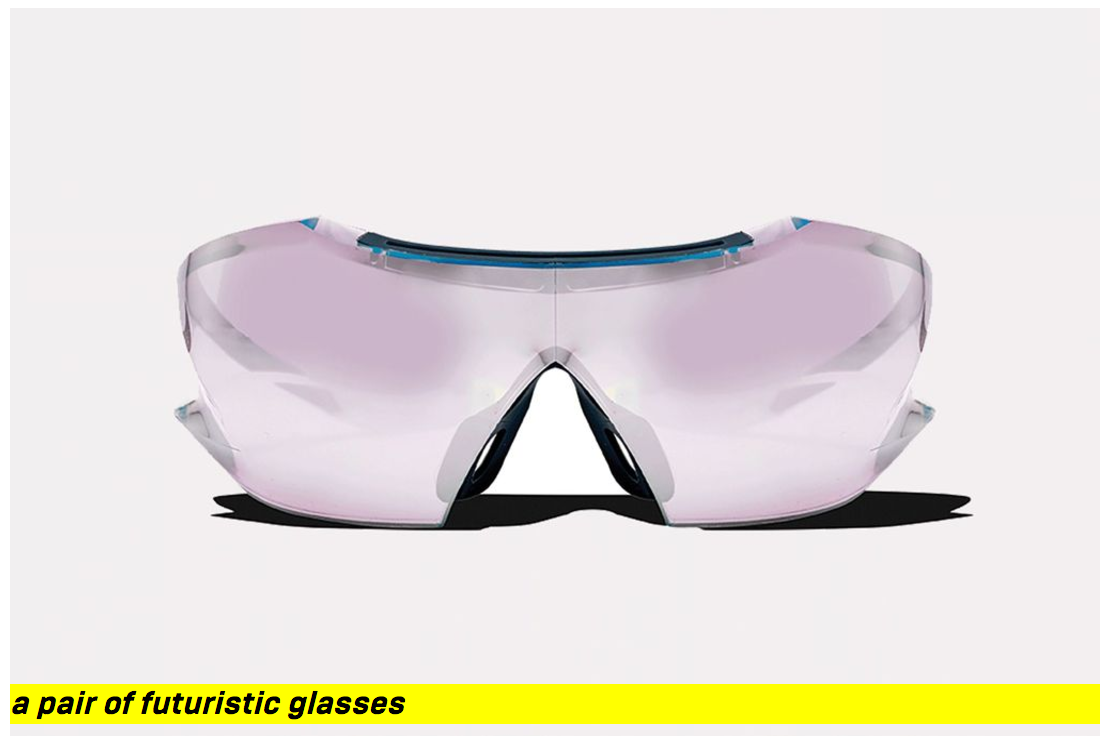}
    \caption{An iconic image from an online magazine captioned by an
      evolved model. The model provides a suitably detailed
      description without any unnecessary context.}
    \label{fg:glasses}
  \end{center}
\end{figure}

Once the model was fully-trained, it was placed on a server where it
can be queried with images to caption. A JavaScript snippet was
written that a developer can embed in his/her site to automatically
query the model to caption all images on a page. This snippet runs in
an existing Chrome extension for custom scripts and automatically
captions images as the user browses the web. These tools add captions
to the `alt' field of images, which screen readers can then read to
blind internet users (Figure~\ref{fg:glasses}).

\subsection{Image Captioning Results}

%

%
%
%

%
%

\begin{figure}[!t]
  \begin{center}
    \includegraphics[width=\columnwidth]{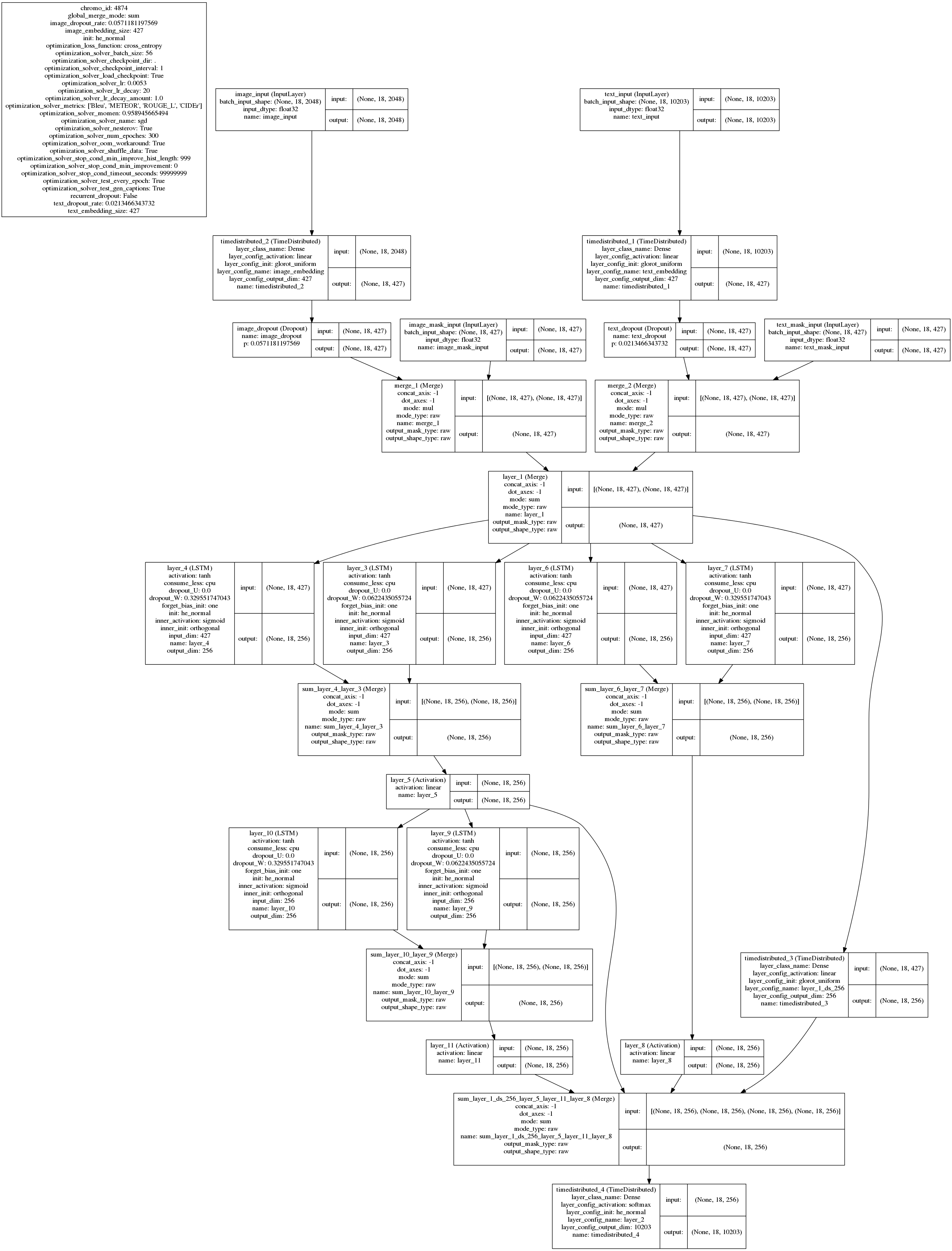}
    \caption{The most fit architecture found by evolution. The full
      image is included in supplementary material. Among the
      components in its unique structure are six LSTM layers, four
      summing merge layers, and several skip connections. A single
      module consisting of two LSTM layers merged by a sum is repeated
      three times. There is a path from the input through dense layers
      to the output that bypasses all LSTM layers, providing the
      softmax with a more direct view of the current input. The motif
      of skip connections with a summing merge is similar to residual
      architectures that are currently popular in deep learning
      \cite{he:arxiv15, szegedy:arxiv16b}.}
    \label{fg:architecture}
  \end{center}
\end{figure}

Trained in parallel on about 100 GPUs, each generation took around one
hour to complete. The most fit architecture was discovered on
generation 37 (Figure~\ref{fg:architecture}).
This architecture performs better than the hand-tuned baseline
\cite{vinyals:cvpr15} when trained on the MSCOCO data alone
(Table~\ref{table:mscoco_results}).
\begin{table}[]
\begin{tabular}{| l | r | r | r |}
\hline
Model & BLEU-4 & CIDEr & METEOR \\ \hline
DNGO \cite{snoek:icml15} & 26.7 & --- & --- \\ \hline
Baseline \cite{vinyals:cvpr15} & 27.7 & 85.5 & 23.7 \\ \hline
Evolved & \textbf{29.1} & \textbf{88.0} & \textbf{23.8} \\ \hline
\end{tabular}
\caption{The evolved network improves over the hand-designed baseline
  when trained on MSCOCO alone.}
\label{table:mscoco_results}
\end{table}

\begin{figure}[]
  \begin{center}
    \includegraphics[width=0.48\columnwidth]{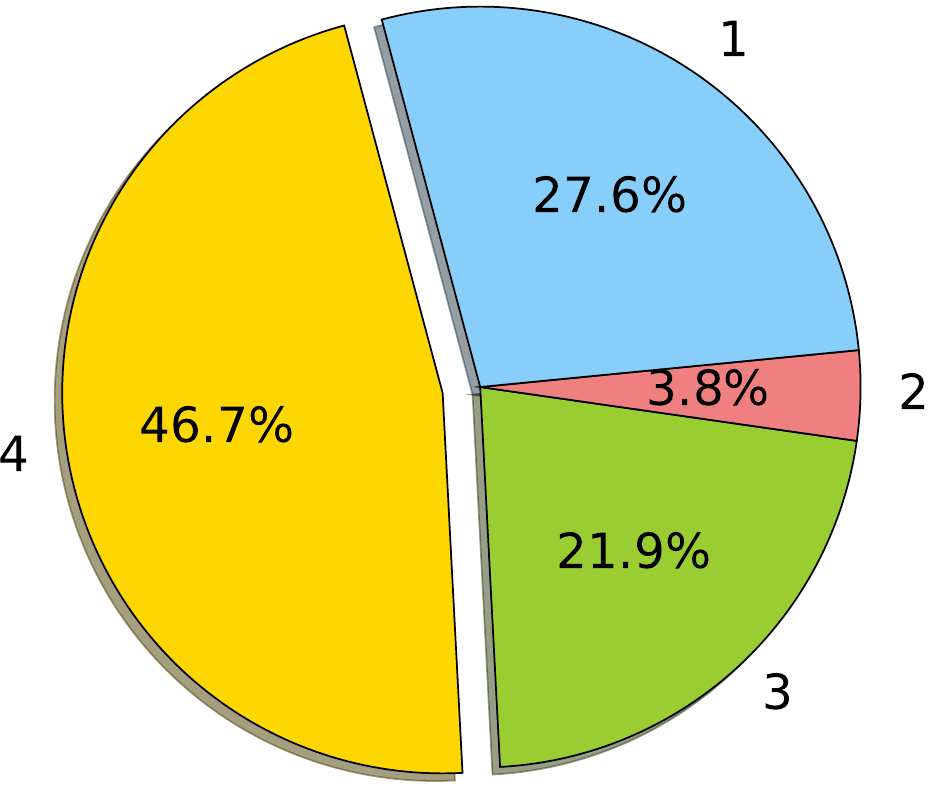}
    \hspace{10pt}
    \includegraphics[width=0.45\columnwidth]{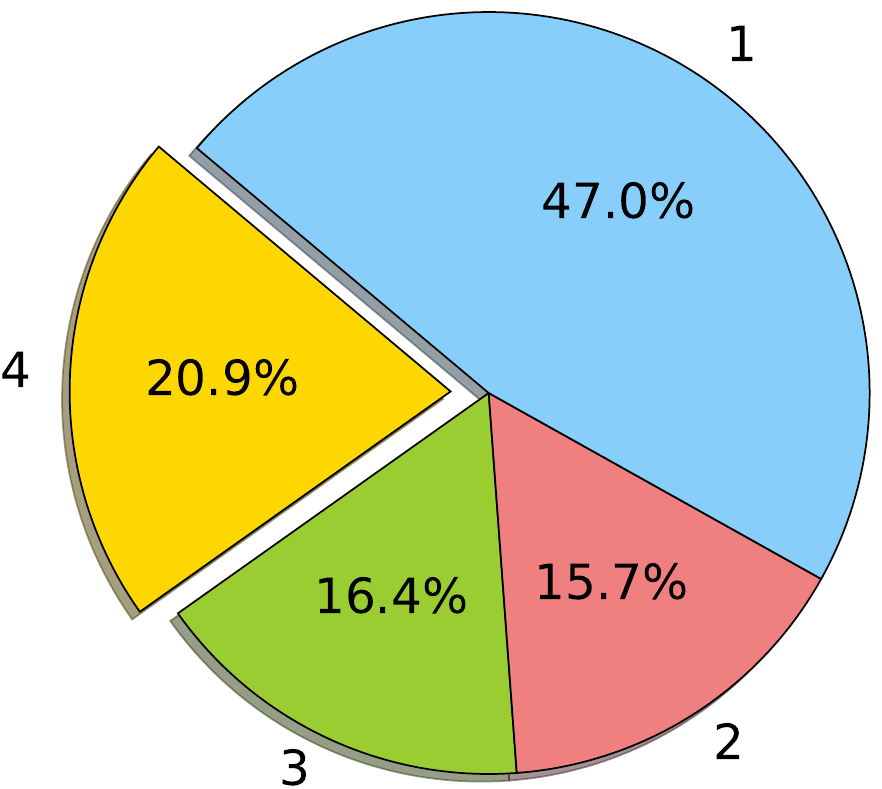}
    \caption{Results for captions generated by an evolved model for
      the online magazine images rated from 1 to 4, with 4=Correct,
      3=Mostly Correct, 2=Mostly Incorrect, 1=Incorrect. Left: On
      iconic images, the model is able to get about one half correct;
      Right: On all images, the model gets about one fifth
      correct. The superior performance on iconic images shows that it
      is useful to build supplementary training sets for specific
      image types.}
    \label{fg:wired_pies}
  \end{center}
\end{figure}

\begin{figure}[]
  \begin{center}
    \includegraphics[width=0.48\textwidth]{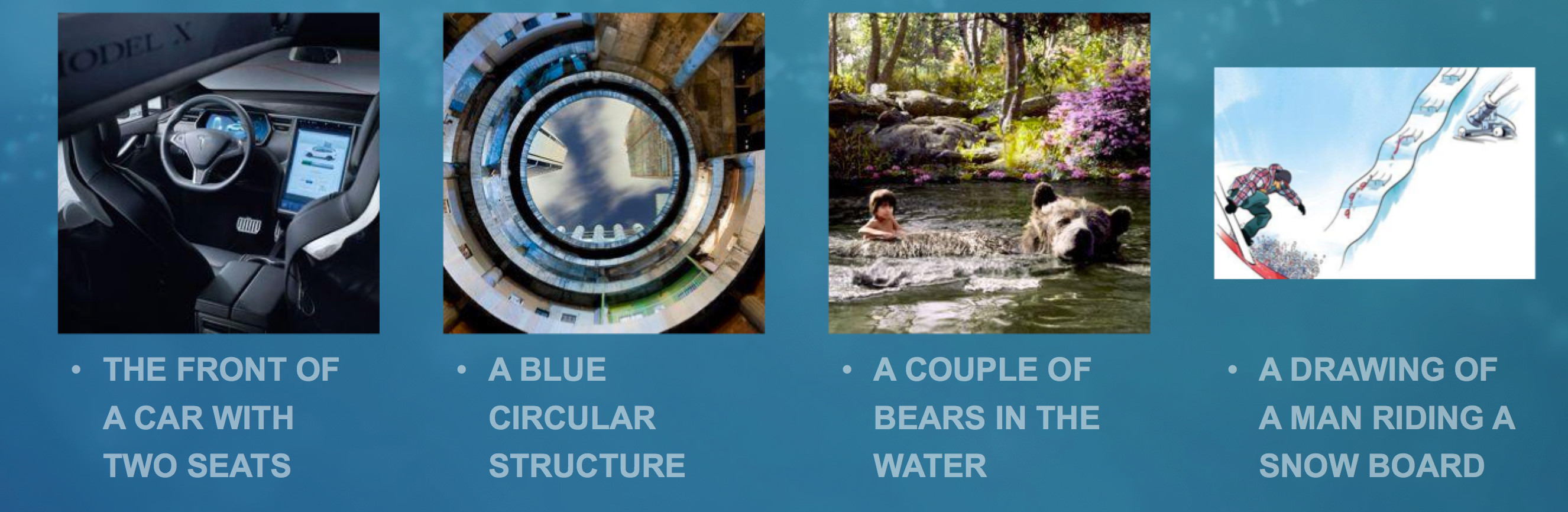}
    \includegraphics[width=0.48\textwidth]{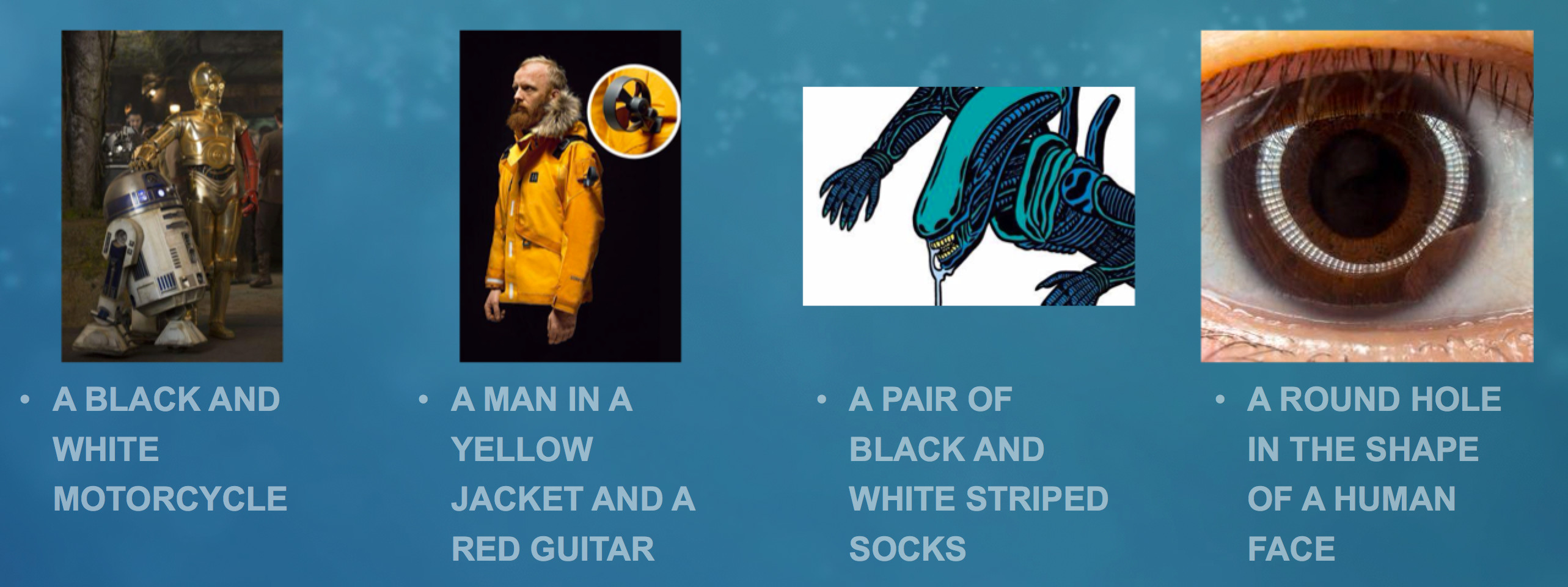}
    \caption{Top: Four good captions. The model is able to abstract
      about ambiguous images and even describe drawings, along with
      photos of objects in context. Bottom: Four bad captions. When it
      fails, the output of the model still contains some correct sense
      of the image.}
    \label{fg:good_and_bad_captions}
  \end{center}
\end{figure}

However, a more important result is the performance of this network on
the magazine website. Because no suitable automatic metrics exist for
the types of captions collected for the magazine website (and existing
metrics are very noisy when there is only one reference caption),
captions generated by the evolved model on all 3100 holdout images
were manually evaluated on a scale from 1 to 4
(Figure~\ref{fg:wired_pies}).
Figure~\ref{fg:good_and_bad_captions} shows some examples of good and
bad captions for these images. 

The model is not perfect, but the results are promising.  There are
many known improvements that can be implemented, including ensembling
diverse architectures generated by evolution, fine-tuning of the
ImageNet model, using a more recent ImageNet model, and performing
beam search or scheduled sampling during training
\cite{vinyals:pami16}.  For this application, it is also important to
include methods for automatically evaluation caption quality and
filtering captions that would give an incorrect impression to a blind
user.  However, even without these additions, the results demonstrate
that it is now possible to develop practical applications through
evolving DNNs.

\section{Discussion and Future Work}
\label{sec:future}

The results in this paper show that the evolutionary approach to
optimizing deep neural networks is feasible: The results are
comparable hand-designed architectures in benchmark tasks, and it is
possible to build real-world applications based on the approach.  It
is important to note that the approach has not yet been pushed to its
full potential. It takes a couple of days to train each deep neural
network on a state-of-the-art GPU, and over the course of evolution,
thousands of them need to be trained. Therefore, the results are
limited by the available computational power.  Interestingly, since it
was necessary to train networks only partially during evolution,
evolution is biased towards discovering fast learners instead of top
performers. This is an interesting result on its own: evolution can be
guided with goals other than simply accuracy, including training time,
execution time, or memory requirements of the network.

Significantly more computational resources are likely to become
available in the near future. Already cloud-based services such as
Amazon Web Services offer GPU computation with a reasonable cost, and
efforts to harness idle cycles on gaming center GPUs are underway.  At
Sentient, for example, a distributed AI computing system called
DarkCycle is being built that currently utilizes 2M CPUs and 5000 GPUs
around the world, resulting in a peak performance of 9 petaflops, on
par with the fastest supercomputers in the world. Not many approaches
can take advantage of such power, but evolution of deep learning
neural networks can. The search space of different components and
topologies can be extended, and more hyperparameters be optimized.
Given the results in this paper, this approach is likely to discover
designs that are superior to those that can be developed by hand
today; it is also likely to make it possible to apply deep learning to
a wider array of tasks and applications in the future.

\section{Conclusion}
\label{sec:conclusion}

Evolutionary optimization makes it possible to construct more complex
deep learning architectures than can be done by hand. The topology,
components, and hyperparameters of the architecture can all be
optimized simultaneously to fit the requirements of the task,
resulting in superior performance. This automated design can make new
applications of deep learning possible in vision, speech, language,
and other areas. Currently such designs are comparable with best human
designs; with anticipated increases in computing power, they should
soon surpass them, putting the power to good use.


\bibliographystyle{ACM-Reference-Format}
\bibliography{/u/nn/bibs/nnstrings,/u/nn/bibs/nn,/u/risto/risto,img_cap,lstm}

\end{document}